%% file: retreg14.tex
\documentclass{article}
\usepackage{fullpage,charter}
\usepackage{graphicx} 
\usepackage{subfigure}

\usepackage{amsmath,amsthm,amssymb} 

\usepackage{algorithm}
\usepackage{algorithmic}

\title{Retrieval of Experiments by Efficient Estimation of\\Marginal Likelihood}

\author{Sohan Seth$^1$, John Shawe-Taylor$^2$, Samuel Kaski$^{1,3}$\\
$^1$Helsinki Institute for Information Technology HIIT,\\
Department of Information of Computer Science,
Aalto University, Finland \\
$^2$Centre for Computational Statistics and Machine Learning,\\
 University College London, UK \\
$^3$Helsinki Institute for Information Technology HIIT,\\
Department of Computer Science, University of Helsinki, Finland\\
$^1$sohan.seth@hiit.fi, $^2$j.shawe-taylor@ucl.ac.uk, $^3$samuel.kaski@hiit.fi}

\newcommand{\E}{\mathbb{E}}
\date{}

\allowdisplaybreaks
\begin{document}
\maketitle 
\begin{abstract}
 
We study the task of retrieving relevant experiments given a query
experiment. By experiment, we mean a collection of measurements from a set of
`covariates' and the associated `outcomes'.  While similar experiments can be
retrieved by comparing available `annotations', this approach ignores the valuable
information available in the measurements themselves. 
To incorporate this information in the retrieval task, we suggest
employing a retrieval metric that utilizes probabilistic models learned from the 
measurements.
We argue that such a metric is a sensible measure of similarity between two
experiments since it permits inclusion of experiment-specific prior knowledge.
However, accurate models are often not analytical, and
one must resort to storing posterior samples which demands 
considerable resources. Therefore, we study strategies to
select informative posterior samples 
to reduce the computational load while maintaining the retrieval performance.
We demonstrate the efficacy of
our approach on simulated data with simple linear regression as the models, and
real world datasets.
\end{abstract}
 
\input{newcommands.tex}

\section{Introduction}

An experiment is an organized procedure for validating a hypothesis, and usually
comprises measurements over a set of variables that are either varied 
(covariates or independent variables) or studied (outcomes or dependent variables).
For example, in the study of genome-wide association, 
one explores the association between `traits' (controlled variable) 
and common genetic variations (response variables) \cite{ge_genetic_2009},
or in the study of functional genomics covariates can be the species, 
disease state, and cell type, whereas outcome can be microarray measurements
\cite{lukk_2010}.  

Traditionally, similar experiments have been retrieved from qualitative
assessment of related scientific documents without explicitly handling the
experimental data.
Recent technological advances have allowed researchers  
to both acquire measurements in an unprecedented scale 
throughout the globe, and to release these measurements 
for public use after curation, e.g., \cite{rustici_arrayexpress_2013}.
However, exploring similar experiments still relies on comparing the manual
annotations which suffer extensively from variations in terminology, and 
incompleteness in annotations, e.g., \cite{baumgartner_manual_2007}.
The global effort of availing researchers with wealth of data invites
the need for sophisticated retrieval systems that look beyond annotations
in comparing related experiments to improve accessibility. 

The next step toward this goal is to compare the \emph{knowledge}
acquired from experimental measurements rather than just annotations.
From a Bayesian perspective, one can quantify knowledge as the
posterior distribution which captures both the information content of
the measurements, in terms of the likelihood function, as well as the
experience and expertise of the experimenter in terms of the prior
distribution.  We explicitly assume that we have access to a database
where researchers have submitted models learned on the experiment
along with measurements and annotations.  We study efficient
approaches for retrieving relevant experiments utilizing this set-up
as a first step toward realizing such an engine.

 We suggest  the conditional
\emph{marginal likelihood}~\eqref{eq:ML}
 as a similarity metric, where the underlying idea is
to evaluate the likelihood of the query experiment on models learned
from (individual) existing experiments.  
Although the
suggested metric can be efficiently estimated as the average posterior
likelihoods over the posterior samples~\eqref{eq:MLavg}, 
this approach has two issues:
storing the posterior samples requires considerable resource, and evaluating
each marginal likelihood can be computationally demanding. This paper deals with
selecting \emph{informative} posterior samples to reduce both storage and
computational requirements while maintaining the retrieval performance. 

\begin{figure}[t]
\centering
\includegraphics[scale=1]{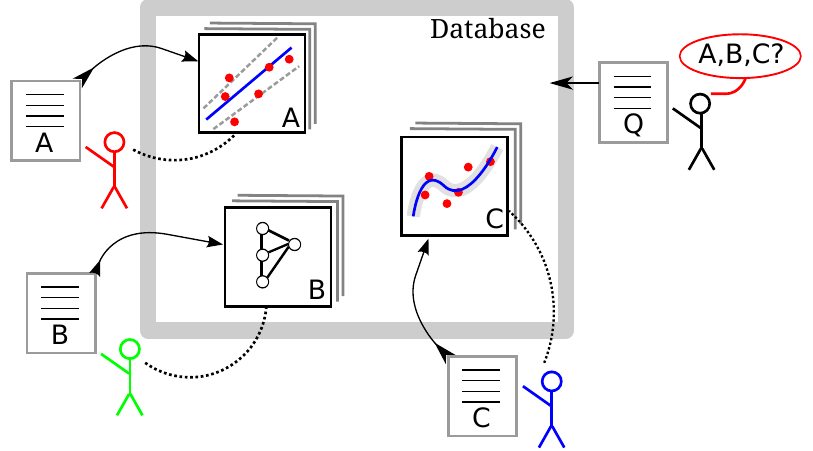}
\caption{
This figure illustrates the basic task we are tackling. Our general objective is to
retrieve experiments A, B or C, given query experiment Q. We achieve this by
measuring similarity between experiments in terms of the marginal likelihood
of the query experiment on the model of the existing experiment. Thus, we
assume that the database contains models of experiments learned by the
experimenter along with the experimental details. Each model is represented in
terms of posterior samples. Our aim is to devise methods to select informative
posterior samples to reduce storage and computational requirements while
preserving the retrieval accuracy. It is to be noted that we can only compare
two experiments if they `share' some common covariates or outcomes.}
\label{fig:retrieval}
\end{figure}

We achieve this by approximating the marginal likelihood as a \emph{weighted
average} of individual likelihoods over posterior samples~\eqref{eq:MLwavg}.
The weights are
then learned to preserve the relative order of experiments in a training set
(section \ref{subsec:rank}).
This is done while imposing 
a suitable sparsity constraint which 
 allows us to only consider posterior samples with non-zero 
weights when computing the likelihood of a query sample, thus reducing the 
storage and computational burden considerably. 
Fig.~\ref{fig:retrieval} illustrates our general objective.

\section{Method} \label{sec:method}

We have a set of experiments $\{\exprm_d\}_{d=1}^D$. Each experiment is defined
as a collection of measurements over covariates and outcomes, i.e.,
$\exprm_d=\{(\b{x}_{di},\b{y}_{di})\}_{i=1}^{n_d}$. We assume that each
experiment $\exprm_d$ has been modeled by a model ${\cal M}_d$, producing a set
of posterior MCMC samples $\{\theta_{dk}\}_{k=1}^{m_d}$ from each model.
Our general objective is to rank the experiments $\exprm_d$---actually the models
$\model_d$ in the database---according to their relevance to a new query
experiment $\exprm_q$ which is not in the database.  

 We
suggest retrieving similar experiments in terms of their marginal likelihood,
\begin{align}\label{eq:ML}
\ML_{q|d}=p(\exprm_q|\exprm_d)
\end{align} 
This metric has been previously discussed in
the context of document retrieval where its use is motivated by capturing
the user's intent in terms of the likelihood of a set of keywords $\exprm_q$
being generated by a document $\exprm_d$ \cite{buntine_scalable_2004}. 
In the context of document
retrieval the marginal likelihood is usually computed by jointly modeling multiple
documents.
However, we cannot evaluate this metric by modeling multiple experiments
jointly, since we explicitly allow experimenters to submit their models.
Therefore, we utilize individual
models $p(\cdot|\exprm_d) \propto p(\exprm_d|\cdot)\pi_d(\cdot)$ to evaluate
the marginal likelihood as $\ML_{q|d} = \E_{p(\cdot|\exprm_d)}p(\exprm_q|\cdot)$,
where $\pi_d$ is the prior information specific to experiment $d$.

\begin{figure}[t]
\centering
\includegraphics[scale=0.9]{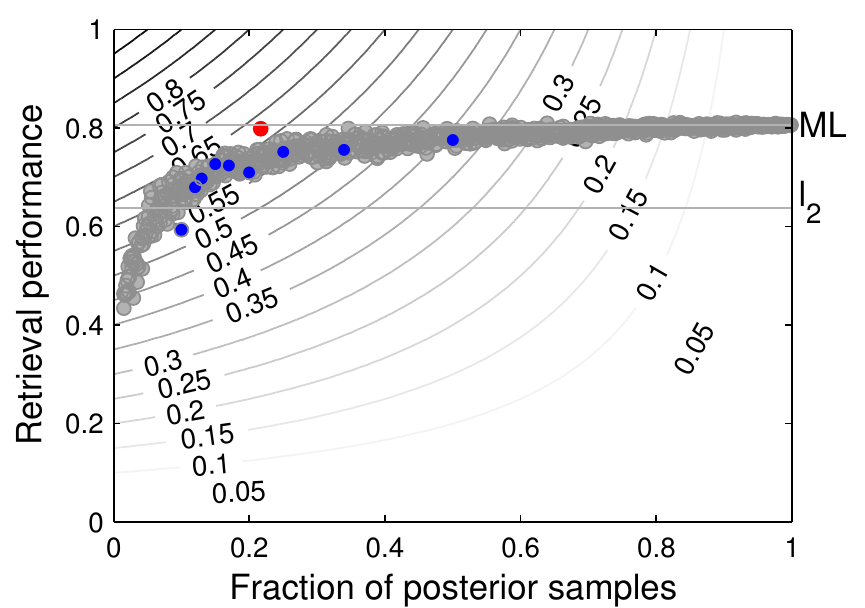}
\caption{The figure illustrates our objective. To evaluate the marginal likelihood ($\ML$) one can 
store every $k$-th posterior sample (blue dots): the choice of $k$ is arbitrary. 
However, this might not be optimal.
For example, selecting posterior samples to be stored randomly (grey dots) might result in better
performance. Our goal is to select informative samples from the pool that
are discriminative between experiments, to reduce computational requirements without
sacrificing retrieval performance (e.g. red dot is achieved by the proposed approach).
It is clear that one encounters a trade-off between the sparsity of the posterior samples,
and the retrieval performance. Therefore, we utilize \emph{(1-sparsity) $\times$ retrieval-performance}
as evaluation metric (contours). In terms of this metric the red dot is close to the  
best blue and grey dots. 
$\ML$ denotes the performance level when all posterior samples are used. 
$l_2$ defines the performance level with $l_2$ distance based metric.}
\label{fig:task}
\end{figure}

The
likelihood can be approximated using posterior samples
$\{\theta_{dk}\}_{k=1}^{m_d} \sim p(\cdot|\exprm_d)$ as 
\begin{align} \label{eq:MLavg}
\widehat{\ML}_{q|d} \approx
\frac{1}{m_d}\sum_{k=1}^{m_d} p(\exprm_q|\theta_{dk}).
\end{align} This approach is 
computationally demanding since one needs to store multiple posterior samples 
$\{\theta_{dk}\}$ and evaluate the corresponding likelihoods $p(\exprm_q|\theta_{dk})$.
The technical contribution of this paper
is to address this issue by selecting \emph{fewer} posterior samples that are essential
in the retrieval task, i.e., discriminative between experiments.
Fig. \ref{fig:task} illustrates our technical 
objective.

 We achieve this by approximating the marginal likelihood as  
\begin{align}\label{eq:MLwavg}
\widetilde{\ML}_{q|d}
\approx \frac{1}{m_d} \sum_{k=1}^{m_d} w_{dk} \prod_{i=1}^{n_d}p((\b{x}_{qi},\b{y}_{qi})|\theta_{dk})
\end{align}
where $\b{w}_d=[w_{d1},\ldots,w_{dm_d}]$ is a vector of \emph{sparse non-negative weights}. 
In this way, the posterior
samples for which the corresponding weights are zero can be safely ignored. Since we are 
effectively estimating the \emph{weighted mean} of a set of values, ideally speaking, $\b{w}_d$
should be a \emph{stochastic} vector: positive values that sum to one. 
However, we observe that even without explicitly imposing this
constraint we can achieve favorable performance, and this simplifies the optimization problem considerably.

\subsection{Preserving ranking of experiments} \label{subsec:rank}
To learn the weights for each experiment, we adapt the concept of
\emph{learning to rank} which is a well explored research problem in 
information retrieval \cite{DBLP:conf/icml/BurgesSRLDHH05}. 
However, while this approach is usually applied for learning a function
over document-query pairs, we utilize the concept in learning weights
over posterior samples for all experiments (``documents'') together.

Assume, without loss of generality, that given a query 
$q$ and two experiments $i_1$ and $i_2$ in the database, $i_1$ ranks higher than
$i_2$, i.e., $\ML_{q|i_1}>\ML_{q|i_2}$.
Therefore, while learning the weights $\b{w}_{i_1}$ and $\b{w}_{i_2}$, we need
to ensure that
\[ \sum_{k}w_{i_1k}p(\expr_q|\theta_{i_1k}) >
\sum_{k}w_{i_2k}p(\expr_q|\theta_{i_2k}).\] When each experiment in
the training set is used as a query $q$, preserving the relative ranks
of each pair $\{i_1,i_2\} \subset \{1,\ldots,D\} \setminus \{q\}$
translates to needing to satisfy $D(D-1)(D-2)$ binary constraints for
learning the weight vectors $\b{w}_1,\ldots,\b{w}_d$.  Fortunately not
all of the constraints are usually required since a user is often
interested in retrieving only the top (say, top $K$) experiments
rather than all experiments.

Therefore, we reformulate our approach and, given a query
$q$, focus on  preserving the order of top $K$ experiments. 
Given any experiment $q$ we select the $K$ closest experiments, $I_q^K =
\{i_{j_1},\ldots,i_{j_K}\}$, and compare them pairwise with the rest of the
$(D-2)$ experiments in the database.  Intuitively, this preserves the relative
orders among the top $K$ experiments $I_q^K$, and also ensures that these
experiments are ranked higher compared to the rest of the $\{1,\ldots,D\}\setminus
\{q \cup I_q^K\}$ experiments. 
This reduces the set
of constraints to $KD(D-2)$ where $K\ll D$.
Notice that it is certainly feasible to choose different $K$ for different queries.

\subsection{Optimization problem} 
\begin{figure*}[t]
\centering
\includegraphics[width=1\textwidth]{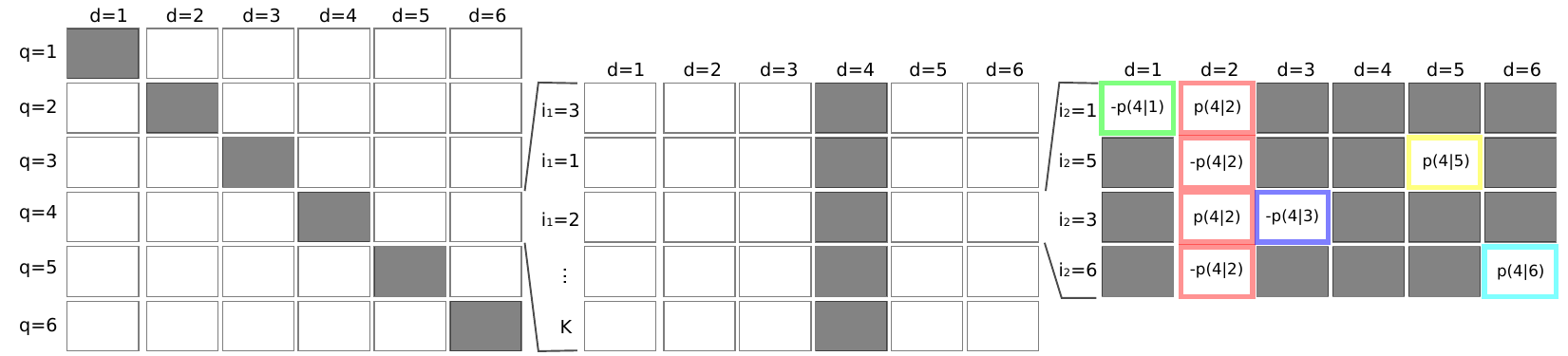}
\caption{Illustration of design matrix  for $D=6$ with notation $p(a|b) = p(\exprm_a|\theta_{b,1:m_b})$.
The matrix is $D(D-2)K\times m$ dimensional where
$m=\sum_d m_d$ is total number of posterior samples from $D$ experiments. 
The second and third figure are zoomed versions of a block of rows of the first and 
second figure respectively. 
Each row of the design matrix belongs to a triplet $(q,i_1,i_2)$, a query and
two retrieved experiments.
The matrix is sparse: each row only has $m_{i_1}+m_{i_2}$ non-zero entries
$\{+p(\exprm_q|\theta_{i_{1}1}),\ldots,
+p(\exprm_q|\theta_{i_{1}m_{i_1}}),
-p(\exprm_q|\theta_{i_{2}1}),\ldots,
-p(\exprm_q|\theta_{i_{2}m_{i_2}})
\}$ corresponding to posterior samples of 
$i_1$ and $i_2$. 
The signs of the entries and corresponding target can be switched arbitrarily. The matrix
contains repeated entries, e.g., the red blocks. We do not
consider $\ML_{q|q}$, so the diagonal blocks of the first figure are zero (gray). 
}
\label{fig:design}
\end{figure*}

Satisfying the binary constraints can be formalized as a classification problem
 $\{ (\b{X}_l, y_l) \}_{l=1}^L$ with a highly sparse design matrix
$\b{X}$ of dimension $L\times m$ (as depicted in Fig.~\ref{fig:design}), 
with $L = KD(D-2)$ realizations and $m = \sum_d m_d$ features 
for learning a combined weight vector $\b{w}=[\b{w}_1,\ldots,\b{w}_d]$, i.e.,
to satisfy $(\b{X}_l\b{w} + b)y_l > 0$ for all $l$.
Each row of $\b{X}$ belongs to a triplet $(q,i_1,i_2)$, and in that
row only the columns associated with  posterior samples from $i_1$ and $i_2$
are non-zero, and have values $\{p(\expr_q|\theta_{i_1k})\}_{k=1}^{m_{i_1}}$ and
$\{-p(\expr_q|\theta_{i_2k})\}_{k=1}^{m_{i_2}}$ respectively. 
The label associated with this entry is 1 if
$\ML_{q|i_1} > \ML_{q|i_2}$, and zero otherwise. An important aspect of this
construction is that the label is not absolute, i.e., we can change the sign 
of a row in the design matrix, i.e., 
assign the values $\{-p(\expr_q|\theta_{i_1k})\}$ and
$\{p(\expr_q|\theta_{i_2k})\}$ to the row instead, and switch the label accordingly. Actually, we
randomly pick one of these scenarios to maintain class
balance, i.e., we have similar numbers of zeros and ones.

Since we are solving a classification problem, each row of the design
matrix can be normalized without effecting the class label. This helps
solve scaling issues: Instead of likelihoods $p_l$, we can classify
log likelihoods $\ln p_l$, and compute the normalized entries as
$\pm\exp(\ln p_l - \max_l\ln p_l)$. These values are in $[-1,1]$.

We use the library \textrm{liblinear}
\cite{Fan:2008:LLL:1390681.1442794} to solve this optimization
problem.  We use the logistic cost with $l_1$ regularization, and set the
regularization parameter to 1. An interesting property of this
approach is that the number of posterior samples with non-zero weights
can be different for different models as needed.

\begin{figure*}[t]
\centering
\includegraphics[width=1\textwidth]{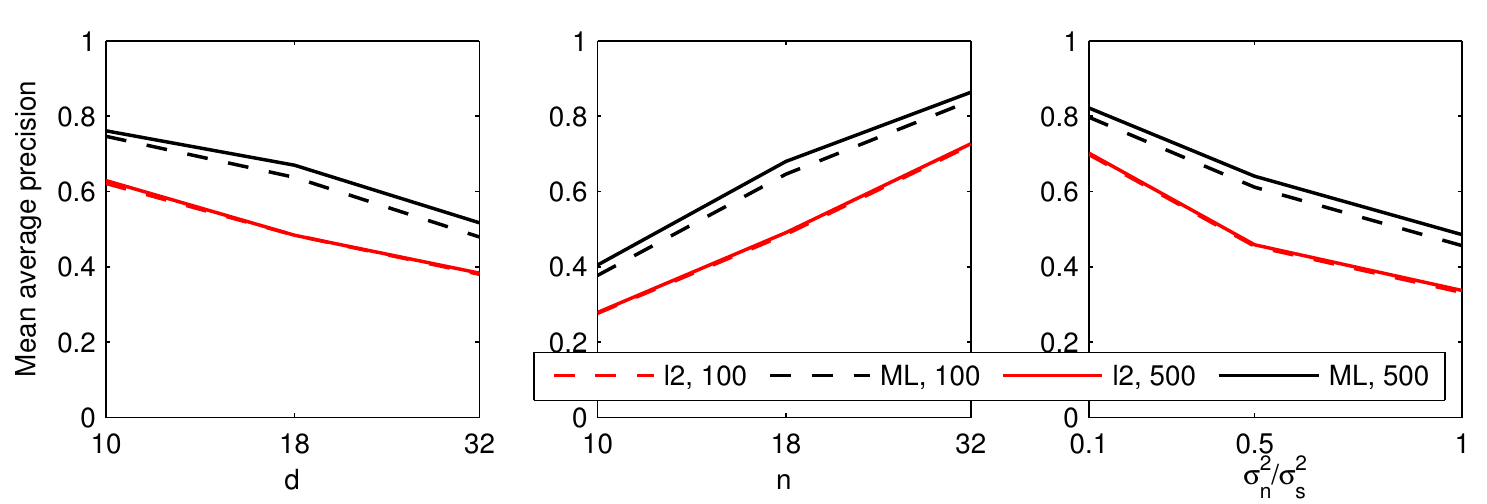}
\caption{Comparison of retrieval performance of the proposed
  probabilistic metric $\ML_{q|d}$ to an ordinary $l_2$ metric
  computed between posterior means of the relation
  $\hat{f}:\b{x}\to\b{y}$. Each experiment has been treated as a
  regression problem with $d$ input features, one output feature, and
  $n$ measurements, i.e., $f \equiv \b{w}$. The plots show the
  variation of mean average precision (MAP) as a function of the
  dimensionality $d$, number of samples $n$ and signal to noise ratio,
  for 100 and 500 posterior samples respectively.  For each plot the
  other two parameters have been averaged over. We observe that the
  probabilistic metric consistently outperforms the  $l_2$
  metric. Also the performance of $\ML_{q|d}$ improves with the number
  of posterior samples. See section~\ref{sec:comparel2} for details.}
\label{fig:compareL2}
\end{figure*}

\begin{figure*}[t]
\centering
\includegraphics[width=1\textwidth]{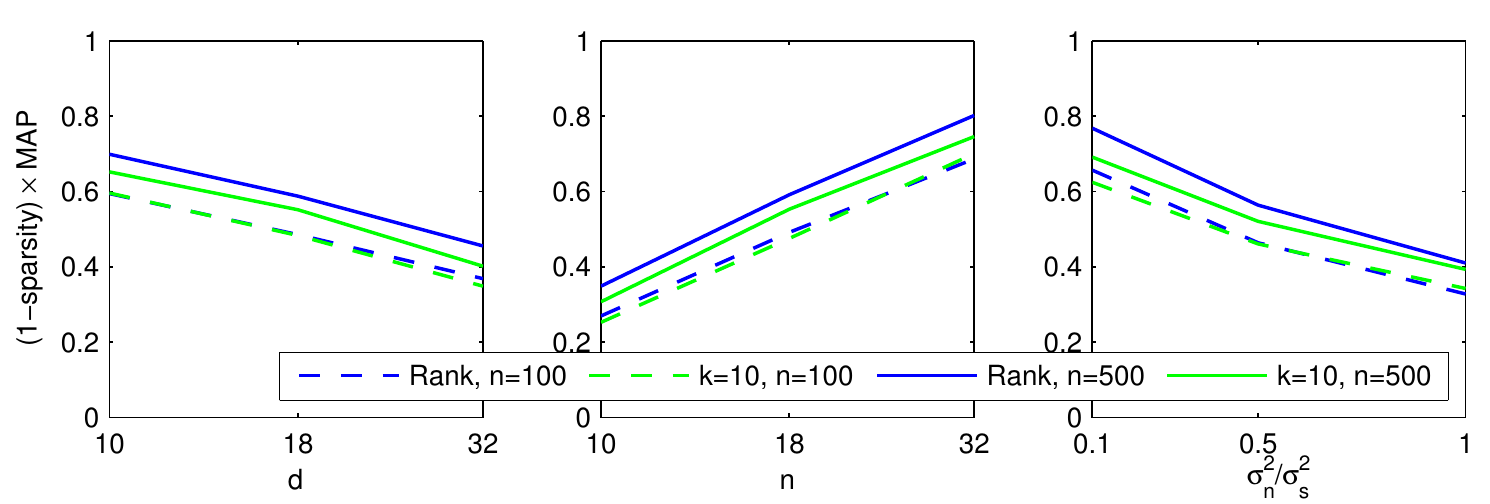}
\caption{Comparison of retrieval performance between probabilistic
  metrics estimated from straight-forwardly picking every $k$-th
  posterior sample, and estimated after learning a weight vector by
  preserving ranks. The evaluation metric focuses on whether the
  method has improved retrieval performance while decreasing the
  number of posterior samples stored. The experiments are simple
  regression tasks $\b{w}:\reals^d\to\reals$ where the ground truth
  regressors come in clusters. The figures show the performance as a
  function of the dimensionality (d), number of measurements (n) and
  signal-to-noise ratio, where the other two features have been
  averaged over. The total posterior samples are either 100 or 500.
  We observe that the proposed ranking-based approach outperforms the
  alternative of storing every $k$-th posterior sample, in particular
  for high signal-to-noise-ratios.  See section~\ref{sec:compareclass}
  for details.}
\label{fig:comparerank}
\end{figure*}

Although we do not restrict the weight vectors to be positive and
normalized to one, we observe that the non-negativity becomes
satisfied naturally, whereas the sum-up-to-one constraint can be
ignored since we are only interested in the ranks.  It is to be noted
that, since we optimize all the weights together, there is a
possibility that all weights from a particular experiment become set
to zero to achieve a sparser solution.  This can happen in particular
when the number of experiments per class is imbalanced, or if an
experiment is an outlier in the sense that it is ranked low most of
the times.  We leave solving this issue, by imposing additional
constraints, for future work.

\section{Related works} \label{sec:approach}

The state-of-the-art in retrieval of experiments is annotation-driven search,
where the user queries with a keyword, and results that match the keyword are returned. 
For example, the experimental factor ontology
\cite{malone_modeling_2010} provides an excellent platform for
retrieving gene expression experiments. However, this approach
requires extensive manual curation to fit the different terminologies
chosen by different groups and researchers, and is obviously not
usable for finding phenomena the experimenter either did not notice or
annotate.

One can take a step further, and compare experiments based on the relation
between covariates and outcomes, i.e., $f:\b{x}\to\b{y}$ with some distance
metric $d(f_i,f_j)$, where $f_i,f_j$ are \emph{point estimates} of the 
relations. If $f$ is linear, $l_2$ can be a suitable distance measure.
It is also possible to explore
the similarity in either the covariates or the outcomes alone in terms of a suitable
representation.
This approach has not been taken in the literature
yet; it has the obvious limitation of not capturing the uncertainty in $f$, and
possible multi-modality of the $p(f)$. 
We empirically demonstrate that capturing this uncertainty 
improves the retrieval performance (section \ref{sec:comparel2}).

We use $\ML_{q|d}$ because it implies a very natural definition of
relevance: experiment $d$ is relevant to $q$ if a model of $q$ would
also be a good model of $d$. Additionally, the definition brings
intrinsic properties not satisfied by most of the other possible
approaches. First, one can compare two experiments that do not share
the same feature space, for instance, one having missing features;
those features can be marginalized out while computing the marginal
likelihood. Second, the models for the existing experiments do not
need to belong to the same family, and one can choose different models
for an experiment as long as the likelihood of the other experiments
can be evaluated in terms of that model.  Third, since each experiment
is modeled separately, the experimenter can include her 
experiment-specific prior knowledge in the model.

One could also consider the similarity $\ML_{d|q} =
p(\exprm_d|\exprm_q)$ which can be evaluated if one has the model of the query
 experiment and the measurements from the previous experiments.  However,
first, this approach would implicitly assume that the querying experimenter is
already capable of modeling the experiment properly, which somewhat contradicts the
purpose of the retrieval. Second, since experiments $\exprm_d$ can have different
number of observations $n_d$, this metric is excessively dependent on number
of observations: small $n_d$ may result in larger likelihood. 

If one models the query experiment as well, then there are other possible
approaches of evaluating similarity between two experiments. For example, \cite{dutta_retrieval_2013}
have recently suggested modeling posterior samples $\{\theta_{dk}\}$ 
sequentially with Dirichlet process mixtures of normal distributions
using particle filtering. Once this model (over posterior samples) has been learned, the
similarity between two experiments can be evaluated through similarity
of the cluster assignments
of the respective posterior samples.
Given models of the query and the existing experiments, one can also evaluate
their similarity in terms of probabilistic distances or kernels
\cite{muandet_learning_2012}.  However, both these approaches have the limitation that
the models have to belong to the same family for the similarity to be defined.
Moreover, the distances or kernels between models are primarily chosen
to satisfy only general properties such as the triangular inequality and positive
definiteness, rather than assisting in the user's task, in our case retrieval.

Another possible approach for measuring similarity between experiments is to
model the measurements together in a multi-task learning framework \cite{Xue_2007}. However, off-the-shelf methods for modeling
multiple experiments together utilize the same prior and likelihood for all experiments which
restricts the generality, and will not exploit the benefit of the
knowledge available at the experimenter's disposal. That said, the true
purpose of multi-task learning is to utilize knowledge from similar tasks to
improve the learning of a new task, which is fundamentally different than retrieval. 
Also, treating each experiment or model separately rather than as part of a unified model
provides well desired modularity to separate the modeling and retrieval task that
can be handled by respective experts.

A similar problem has been explored before by \cite{caldas_probabilistic_2009} where the authors
aimed at retrieving a single sample given a query sample.  This was done
by modeling multiple samples together using latent Dirichlet allocation.
Retrieving an experiment given a query experiment, however, is
conceptually very different since a single sample cannot capture the
experimental variability that one might be interested in.  That said, retrieval of
experiments as discussed in this article allows one to also query with a single
observation to find the closest experiment which could have generated
that particular
sample.  This approach has an 
intriguing characteristic that it enables assigning different parts
of the query experiment to different models.

\section{Simulations} \label{sec:results} 
We study the performance and features of the proposed
approaches in a simple set-up where the relation between covariates 
$\b{x}\in\reals^d$ and outcome $y$
is assumed to be linear, and corrupted by additive Gaussian noise. Thus each
experiment $\exprm_i$ can be described by the linear relation 
$\b{w}_i:y = \b{w}_i^\top\b{x}+\epsilon$.
In order to create a \emph{ground truth}, the experiments are assumed
to come in clusters,
where each cluster is centered at $\b{w}^\ast_i$, $i=1,\ldots,C$,
where $C$ is the number of clusters.
Thus each retrieved experiment can be classified as either relevant or irrelevant
depending on whether it shares the same cluster with the query, 
and the retrieval performance can
be evaluated using a standard metric such as \emph{mean average
  precision} (MAP) \cite{Smucker:2007}.
For a fair comparison, we do not use any experiment-specific prior information
during modeling since our objective here is to discuss that, first, the
proposed retrieval metric performs reasonably well compared to trivial
retrieval metrics, and second, rank preservation leads to
similar retrieval performance using only a fraction of posterior samples. 

We generate experiments $\exprm_i \equiv \b{w}_i$ 
from $C=20$ clusters. The number of experiments within each cluster is generated from a
Poisson distribution with rate $10$, thus, we have $\sim 200$ experiments. 
We randomly split the 
experiments in two groups: $75\%$ of the experiments are treated as
the database
and used for training, and the rest are used as queries.
The number of measurements in each experiment is chosen from a Poisson 
distribution with rate $n$, and we generate $m$ posterior samples
from a regression model with Gaussian likelihood and sparse gamma prior 
over the weight precisions. We use JAGS to generate the posterior samples. 
To evaluate the
performance over different parameter settings we choose $d\in\{10,18,32\}$, $n
\in \{10,18,32\}$, $m\in\{100,500\}$, and $\sigma_n^2/\sigma_s^2 \in \{0.1,0.5,1\}$,
thus, 54 set-ups in total.

\subsection{Comparison between $\ML_{q|d}$ and $l_2(\hat{\b{w}}_d,\hat{\b{w}}_q)$} \label{sec:comparel2}
We start by comparing the performance of the marginal likelihood metric over 
the straightforward $l_2$ metric (Fig. \ref{fig:compareL2}). Marginal likelihood consistently outperforms
the ordinary distance between posterior means $\hat{\b{w}}_i$. Here
we have used all posterior samples for computing $\ML_{q|d}$. Notice that we
can easily consider a multimodal posterior distribution where the posterior
mean is not a sufficient descriptor, and this would result in poor retrieval 
performance for the alternative method. Our goal here was to
show that even in simple cases, learning the posterior distribution can assist in
improving the performance. 
\begin{figure*}[t]
\centering
\includegraphics[width=1\textwidth]{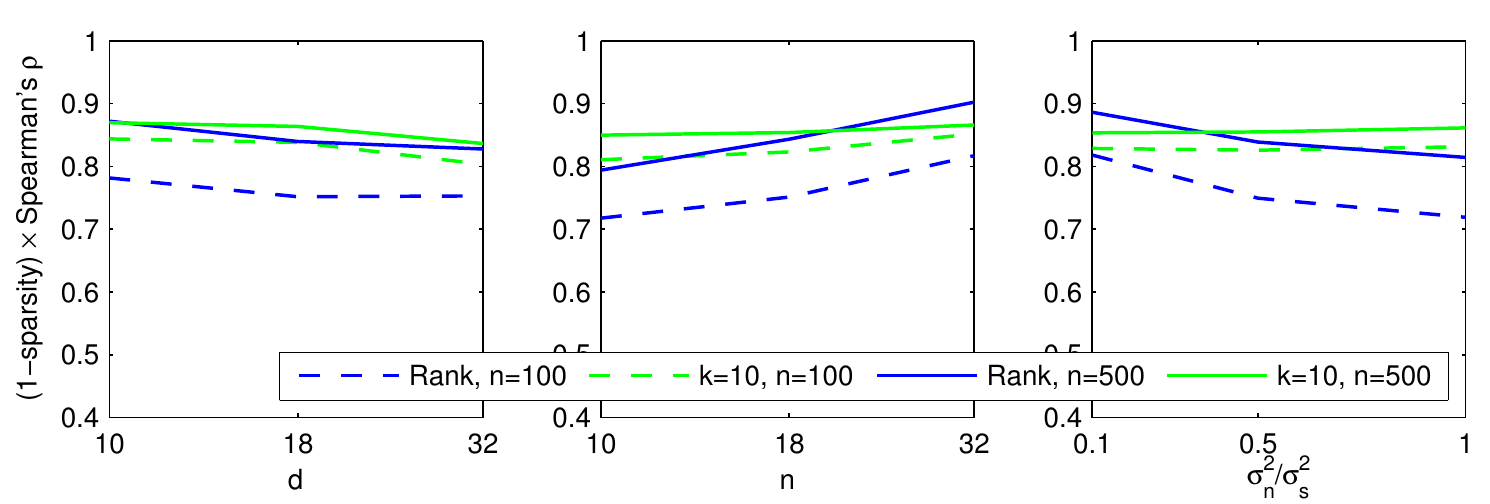}
\caption{Comparison of retrieval performance between probabilistic metrics
estimated by straight-forwardly picking every $k$-th posterior sample, and estimated after
learning a weight vector that preserves ranks. The evaluation metric focuses on
whether the method has improved retrieval performance while decreasing the 
number of posterior samples stored. The experiments are simple regression 
tasks $\b{w}:\reals^d\to\reals$ not forming any clusters. 
The figures show the variation of 
performance as a function of the dimensionality (d), number of measurements 
(n) and signal-to-noise ratio, where the other two features have been averaged
over. The number of posterior samples are either 100 or 500. 
We observe that the proposed approach is generally not better than storing
arbitrary posterior samples. However, as more posterior samples
are given the proposed approach soon catches up. See section~\ref{sec:comparenoclass} for details.}
\label{fig:comparenoclass}
\end{figure*}

\subsection{Comparison between $\widehat{\ML}_{q|d}$ and $\widetilde{\ML}(q|d)$: a representative training set}
\label{sec:compareclass}
We use the same dataset to compare performance between the proposed approach
for reducing the posterior samples to be stored with weighted average of 
likelihood $\widetilde{\ML}_{q|d}$, and a simpler method. To recapitulate, our goal has
been to select from a pool of posterior samples informative ones that can maintain the
retrieval performance. Thus the performance should be better than by simply
storing every $k$-th posterior sample without
any optimization, $\widehat{\ML}_{q|d}$. A small $k$ would improve
computational time but degrade sparsity. 
We compare the performance of the proposed approach against performance with 
$k=10$ in Fig. \ref{fig:comparerank}. Since our objective is to
impose sparsity in the weight vector while improving retrieval performance, we 
evaluate the performance in terms of (1 - sparsity)$\times$
minimum-average-precision (see Fig.~\ref{fig:task}). 
For the rank preservation approach, we present the result
for $K=25$ since the others $K=5,10,15,20$ perform equally well. Therefore, we conclude that
in the presence of representative experiments in the training set, the proposed
approach can safely select informative posterior samples.

\subsection{Comparison between $\widehat{\ML}_{q|d}$ and
  $\widetilde{\ML}(q|d)$: training set not representative}
\label{sec:comparenoclass}

In the previous two sections, we have presented results for the case when 
the experiments come in clusters, and the query belongs to one of them
as well. Intuitively this is a simpler problem since a query always has 
certain representative experiments in the training set. To elaborate,
one can learn the weights for an experiment by preserving 
ranks within the same cluster, and since the query is from one of the clusters, the
learned weight can be used to compute the likelihood of a query reliably. To make
the problem difficult we consider the situation when the experiments
do not have clustered structure. To investigate if the proposed method still performs well in
this `extreme' set-up, we now generate $200$ experiments in the same way but without splitting them
in clusters. Since now we do not have
any ground truth, we consider the ranking given by $\widehat{\ML}_{q|d}$ with all posterior
samples as the ground truth, and evaluate the performance in terms 
of Spearman's correlation with the  ground truth ranking. We observe
(Fig. \ref{fig:comparenoclass}) that when the total number of
posterior  
samples $m$ is low, storing every $k$-th sample performs better. However, as
more posterior samples are added, the proposed method performs equally well.
This situation is analogous to the general finding that  
generalizing beyond the learning data is difficult.

\begin{figure*}[t]
\centering
\includegraphics[width=0.85\textwidth]{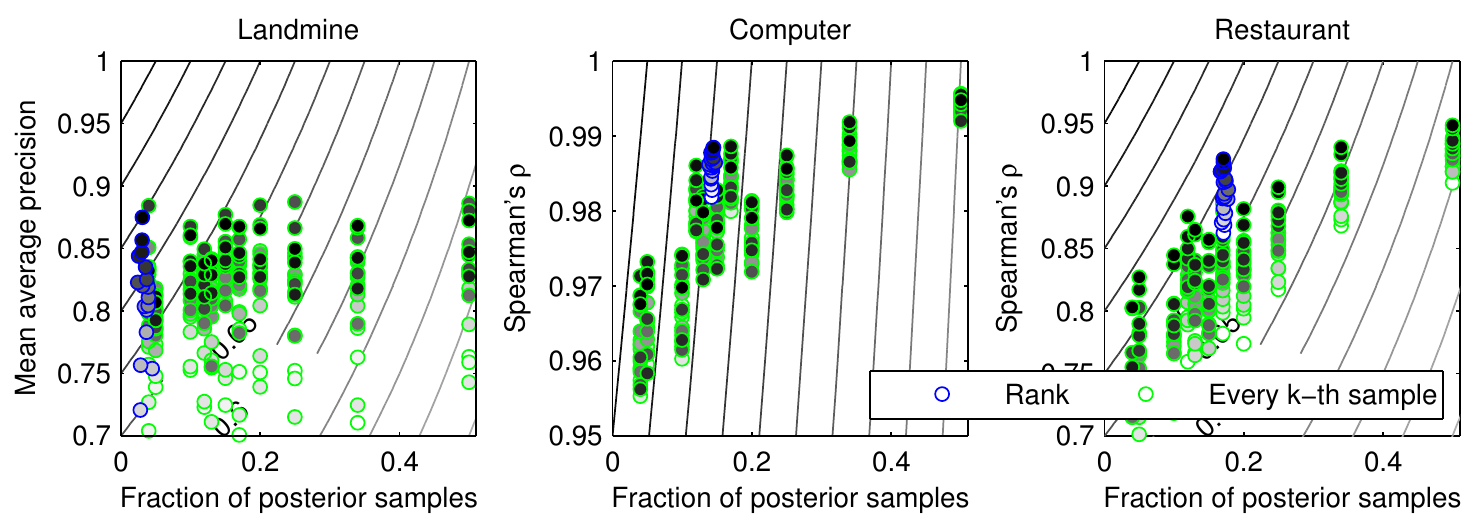}
\caption{Comparison of the proposed approach and a simpler metric on real datasets.
For landmine we present mean average precision MAP as have access to
labels of each experiment, while for the other two datasets we
present the performance compared to $\widehat{\ML}_{q|d}$ estimated with 
all posterior samples.
Each gray shade corresponds to a random partition of the dataset in 
database and queries. The proposed approach shows improved performance
compared to storing every $k$-th sample since its performance is
toward the upper-left corner. The contours are for (1-sparsity) x
retrieval-performance (see 
Fig.~\protect\ref{fig:task}).}
\label{fig:real}
\end{figure*}

\section{Experiments}
We demonstrate the performance of the proposed approach on three real world
datasets: landmine  \cite{Xue_2007}, 
computer \cite{NIPS2006_251}, and restaurant \cite{govea2011}.
The first two are standard in the multi-task learning framework. 
For landmine, we have access
to class labels of each experiment, and we evaluate the performance of our
approach in terms of mean average precision MAP, while on the other two datasets we use correlation
with respect to the ranking given by $\widehat{\ML}_{q|d}$ with all posterior samples.
We present the results collectively in Fig.~\ref{fig:real}.
For landmine, we train a binary probit regression model, while for the
other datasets we use a normal regression model with non-sparse gamma prior
over the weight precisions. For
each experiment we generate 100 posterior samples. For each dataset
we randomly split it 3:1 into the database and queries.

\subsection{Landmine}
The data consist of 29 experiments: each experiment is a
classification task for detecting the presence of either landmine (1) or
clutter (0) from 9 input features. Each experiment has been collected from
either a highly foliated region or a desert-like region. Thus they can be split
in two classes (16-13). We observe that this is a relatively simple problem in the sense that
the classes are well separated, and thus a few posterior samples are
sufficient for good retrieval performance. Due to the same reason, the 
proposed approach is able to retain the retrieval performance using
only very few posterior samples. 

\subsection{Computer}
The data consist of 200 experiments: each experiment is a prediction
task of how a student rates 20 computers in the scale 0-10. Each
computer is described in 13 binary features.  Thus, each experiment
$\reals^{13} \to \reals$ has about 20 samples (some entries missing).
Since there are no obvious ground truth labels, we measure how well
the proposed approach can reduce the number of posterior samples while
preserving rankings. We observe that the problem is relatively simple
since even a few posterior samples have been able to preserve the
ranking with respect to $\widehat{\ML}_{q|d}$. However, the number of
samples stored is larger than in the previous example since there is
no clear clustering.

\subsection{Restaurant}
The data consist of 119 experiments: each experiment is a 
prediction task of how a customer rates 130 restaurants in the scale 1-3. 
All customers
do not rate all available restaurants, and so the number of observations
in each experiments varies, from 3-18. We select 7 categorical features
for each experiment and binarize them, resulting in a $\reals^{22} \to \reals$
regression problem. We observe that this problem is more difficult in the
sense that performance drops when the number of samples is decreased. However,
the proposed approach has been able to collect essential samples to
preserve the true rank better. 

\section{Conclusion and future work}

In this paper we have explored the task of retrieving relevant 
experiments given a query experiment. 
The state of the art is to retrieve by matching textual (categorical)
annotations. We argued that this approach 
is not optimal since it ignores
the actual measurements collected within the experiment, and suggested retrieving experiments
based on the relation between covariates and outcomes that is learned from 
the measurements. However, rather than using a single instance of this relation,
we showed that it is better to model its posterior distribution.
We used a retrieval metric that computes the 
marginal likelihood of a query experiment on the models
learned on the measurements from existing experiments. 

This paper is intended to be a proof of concept towards a potentially
highly useful community effort of extending experiment databanks to
include also knowledge of the experimenters in a rigorously reusable
form, as models. As of now, this is highly non-standard yet would be
beneficial since the experimenter alone is best acquainted with
his/her measurements and is able to train the most sensible model by
incorporating his/her experience as prior knowledge.  Storing models
of experiments can, however, be cumbersome since most often they are
not expressed in an analytic form. A widely applicable alternative is
to store samples of the posterior; we suggested approaches to select
the most informative posterior samples to store.  Notice that
posterior samples can be generated also when one has an analytic
posterior.  We have presented a set of convincing results on simulated
data with regression as a task, as well as on standard real datasets.

\section*{Acknowledgments}
This project is partly supported by the Academy of Finland (Finnish Centre of Excellence in Computational Inference Research COIN, 251170), 
and the Aalto University MIDE (Multidisciplinary Institute of Digitalisation and Energy) research programme. 
The calculations presented above were performed using computer resources within the Aalto University School of Science ``Science-IT'' project.
 
\bibliography{../bib}
\bibliographystyle{unsrt}

\end{document}

%% file: newcommands.tex
\newcommand{\model}{\mathcal{M}}    
\newcommand{\bs}[1]{\boldsymbol{#1}}
\renewcommand{\b}[1]{\textbf{#1}}
\newcommand{\normal}{\mathcal{N}}
\newcommand{\exprm}{\mathcal{E}}
\newcommand{\dist}{\mathrm{dist}}
\newcommand{\expr}{\mathcal{E}}
\newcommand{\data}{\mathcal{D}}
\newcommand{\process}{\mathcal{G}}
\newcommand{\prob}{\mathrm{Pr}}
\newcommand{\invwish}{\mathcal{IW}}
\newcommand{\reals}{\mathbb{R}}
\newcommand{\DP}{\mathrm{DP}}
\newcommand{\PL}{\mathrm{PL}}
\newcommand{\ML}{\mathrm{ML}}